\documentclass[10pt,twocolumn,letterpaper]{article}

\usepackage{cvpr}
\usepackage{times}
\usepackage{epsfig}
\usepackage{graphicx}
\usepackage{amsmath}
\usepackage{amssymb}

\usepackage{cite}
\usepackage{wrapfig}
\usepackage{multirow}
\usepackage{caption}
\usepackage{subcaption}
\usepackage{verbatim}
\usepackage{booktabs}
\usepackage[ruled,vlined,linesnumbered]{algorithm2e}
\usepackage{comment}
\usepackage{amsmath}

\makeatletter
\newcommand*{\rom}[1]{\expandafter\@slowromancap\romannumeral #1@}
\makeatother

\usepackage[pagebackref=true,breaklinks=true,letterpaper=true,colorlinks,bookmarks=false]{hyperref}
\usepackage{amsmath}
\def\cvprPaperID{496} 

\def\eqnvspace{{\vspace{-2mm}}}
\def\figvspace{{\vspace{-5mm}}}

\newcommand{\Paragraph}[1]{\vspace{-0mm} \noindent \textbf{#1} \hspace{0mm}}
\newcommand{\Section}[1]{\vspace{-1mm} \section{#1} \vspace{-1mm}}
\newcommand{\SubSection}[1]{\vspace{-1mm} \subsection{#1} \vspace{-1mm}}
\newcommand{\SubSubSection}[1]{\vspace{-3mm} \subsubsection{#1} \vspace{-1mm}}

\newcommand{\norm}[1]{\left\lVert#1\right\rVert}
\DeclareMathOperator*{\argmax}{arg\,max}

\cvprfinalcopy 

\ifcvprfinal\fi
\begin{document}

\title{Deep Tree Learning for Zero-shot Face Anti-Spoofing}

\author{Yaojie Liu,  Joel Stehouwer,  Amin Jourabloo,  Xiaoming Liu \\
Department of Computer Science and Engineering \\
Michigan State University, East Lansing MI 48824\\
{\tt \{liuyaoj1, stay.jb, jourablo, liuxm\}@msu.edu}
}

\maketitle

\begin{abstract}
Face anti-spoofing is designed to prevent face recognition systems from recognizing fake faces as the genuine users.
While advanced face anti-spoofing methods are developed, new types of spoof attacks are also being created and becoming a threat to all existing systems.
We define the detection of unknown spoof attacks as Zero-Shot Face Anti-spoofing (ZSFA).
Previous ZSFA works only study $1$-$2$ types of spoof attacks, such as print/replay, which limits the insight of this problem.
In this work, we investigate the ZSFA problem in a wide range of $13$ types of spoof attacks, including print, replay, $3$D mask, and so on.
A novel Deep Tree Network (DTN) is proposed to 
partition the spoof samples into semantic sub-groups in an unsupervised fashion.
When a data sample arrives, being know or unknown attacks, DTN routes it to the most similar spoof cluster, and makes the binary decision. 
In addition, to enable the study of ZSFA, we introduce the first face anti-spoofing database that contains diverse types of spoof attacks. 
Experiments show that our proposed method achieves the state of the art on multiple testing protocols of ZSFA.
\end{abstract}

\Section{Introduction}
\label{sec:intro}

Face is one of the most popular biometric modalities  due to its convenience of usage, e.g., access control, phone unlock.  
Despite the high recognition accuracy, face recognition systems are not able to distinguish between real human faces and fake ones, e.g., photograph, screen. 
Thus, they are vulnerable to face spoof attacks, which deceives the systems to recognize as another person.
To safely use face recognition, face anti-spoofing techniques are required to detect spoof attacks before performing recognition.

Attackers can utilize a wide variety of mediums to launch spoof attacks. 
The most common ones are replaying videos/images on digital screens, i.e., replay attack, and printed photograph, i.e., print attack. 
Different methods are proposed to handle replay and print attacks, based on either handcrafted features~\cite{maatta2011face,patel2016secure,boulkenafet2015face} or CNN-based features~\cite{face-anti-spoofing-using-patch-and-depth-based-cnns,Jourabloo_2018_ECCV,learning-deep-models-for-face-anti-spoofing-binary-or-auxiliary-supervision, feng2016integration}.
Recently, high-quality $3$D custom mask is also used for attacking, i.e., $3$D mask attack. 
In~\cite{liu20163d,liu20163d2,Liu_2018_ECCV}, methods for detecting print/replay attacks are found to be less effective for this new spoof, and hence the authors leverage the remote photoplethysmography (r-PPG) to detect the heart rate pulse as the spoofing cue. 
Further, facial makeup may also influence the outcome of recognition, i.e., makeup attack~\cite{chen2013automatic}. 
Many works~\cite{Chang_2018_CVPR,chen2013automatic,chen2014impact} study facial makeup, despite not as an anti-spoofing problem.

\begin{figure}[t!]
\centering
\includegraphics[width=0.9\linewidth]{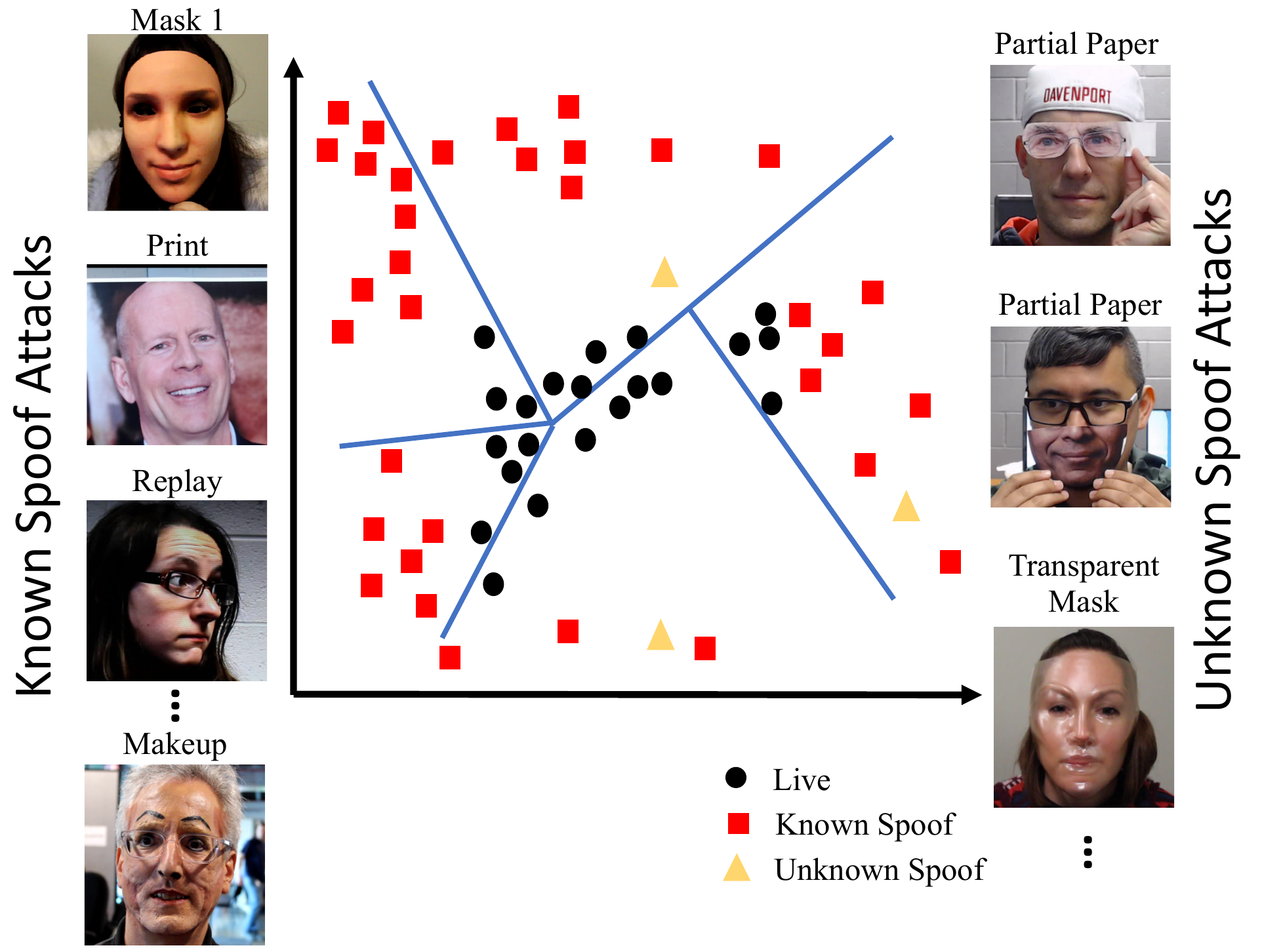}
\vspace{-2mm}
\caption{\small To detect unknown spoof attacks, we propose a Deep Tree Network (DTN) to unsupervisely learn a hierarchic embedding for known spoof attacks. Samples of unknown attacks will be routed through DTN and classified at the destined leaf node.}
\label{fig:concept}
\figvspace 
\end{figure}

All aforementioned methods present algorithmic solutions to the \textit{known} spoof attack(s), where models are trained and tested on the {\it same} type(s) of spoof attacks.
However, in real-world applications, attackers can also initiate spoof attacks that we, the algorithm designers, are not aware of, termed \textit{unknown} spoof attacks\footnote{There is subtle distinction between 1) {\it unseen attacks}, attack types that are known to algorithm designers so that algorithms could be tailored to them, but their data are unseen during training; 2) {\it unknown attacks}, attack types that are neither known to designers nor seen during training. We do not differentiate these two cases and term both unknown attacks.}. 
Researchers increasingly pay attention to the generalization of anti-spoofing models, i.e., how well they are able to detect spoof attacks that have never been seen during the training?
We define the problem of detecting unknown face spoof attacks as \textbf{Zero-Shot Face Anti-spoofing (ZSFA)}.
Despite the success of face anti-spoofing on known attacks, ZSFA, on the other hand, is a new and unsolved challenge to the community. 

The first attempts on ZSFA are~\cite{face-upad,arashloo2017anomaly}. 
They address ZSFA between print and replay attacks, and regard it as an outlier detection problem for live faces (a.k.a.~real human faces). 
With handcrafted features, the live faces are modeled via standard generative models, e.g., GMM, auto-encoder. 
During testing, an unknown attack is detected if it lies outside the estimated live distribution.
These ZSFA works have  three drawbacks:

\Paragraph{Lacking spoof type variety:} Prior models are developed w.r.t.~print and replay attacks only. The respective feature design may not be applicable to different unknown attacks. 

\Paragraph{No spoof knowledge:} Prior models only use live faces, without leveraging the available known spoof data. 
While the unknown attacks are different, the known spoof attacks may still provide valuable information to learn the model. 

\Paragraph{Limitation of feature selection:} They use handcrafted features such as LBP to represent live faces, which were shown to be less effective for known spoof detection~\cite{learning-deep-models-for-face-anti-spoofing-binary-or-auxiliary-supervision,li2016original, patel2016cross, yang2014learn}. 
Recent deep learning models~\cite{learning-deep-models-for-face-anti-spoofing-binary-or-auxiliary-supervision,Jourabloo_2018_ECCV} show the advantage of CNN models for face anti-spoofing.

This work aims to address all three drawbacks.
Since one ZSFA model may perform differently when the unknown spoof attack is different, it should be evaluated on a wide range of unknown attacks types. 
In this work, we substantially expand the study of ZSFA from $2$ types of spoof attacks to $13$ types. 
Besides print and replay attacks, we include $5$ types of $3$D mask attacks, $3$ types of makeup attacks, and $3$ partial attacks. 
These attacks cover both impersonation spoofing, i.e., attempt to be authenticated as someone else, and obfuscation spoofing, i.e., attempt to cover attacker's own identity. 
We collect the first face anti-spoofing database that includes these diverse spoof attacks, termed Spoof in the Wild database with Multiple Attack Types (SiW-M).

To tackle the broader ZSFA, we propose a Deep Tree Network (DTN). 
Assuming there are both homogeneous features among different spoof types and distinct features within each spoof type, a tree-like model is well-suited to handle this case: learning the homogeneous features in the early tree nodes and distinct features in later tree nodes.
Without any auxiliary labels of spoof types, DTN learns to partition data in an unsupervised manner. 
At each tree node, the partition is performed along the direction of the largest data variation.
In the end, it clusters the data into several sub-groups at the leaf level, and learns to detect spoof attacks for each sub-group independently, shown in Fig.~\ref{fig:concept}. 
During the testing, a data sample is routed to the most similar leaf node to produce a binary decision of live vs.~spoof.

In summary, our contributions in this work include :

$\bullet$ Conduct an extensive study of zero-shot face anti-spoofing on $13$ different types of spoof attacks;

$\bullet$ Propose a Deep Tree Network (DTN) to learn features hierarchically and detect unknown spoof attacks;

$\bullet$ Collect a new database for ZSFA and achieve the state-of-the-art performance on multiple testing protocols.

\Section{Prior Work}
\label{sec:prior}
 
\begin{table*}[t!]
\small
	\centering
	\caption{Comparing our SiW-M with existing face anti-spoofing datasets.}
\vspace{-3mm}
	\resizebox{\textwidth}{!} 
{
	\begin{tabular}{l|c|c|c|c|c|c|c|c|c|c|c}
		\hline
	\multirow{2}{*}{Dataset} & \multirow{2}{*}{Year}& Num. of & \multicolumn{3}{c|}{Face variations} & \multicolumn{5}{c|}{Spoof attack types} & Total num. of\\ \cline{4-11}
&& subj./vid. & pose & expression & lighting & replay & print & $3$D mask & makeup & partial & spoof types\\ \hline
			
CASIA-FASD~\cite{zhang2012face}&$2012$& $50$/$600$& Frontal & No&No& $1$& $2$& $0$ & $0$ & $0$ & $3$  \\ \hline

Replay-Attack~\cite{Chingovska_BIOSIG_2012}&$2012$&$50$/$1,200$& Frontal &No &  Yes&$1$&$1$& $0$ & $0$ & $0$ & $2$\\ \hline

HKBU-MARs~\cite{liu20163d}&$2016$&$35$/$1,008$& Frontal& No &Yes&$0$&$0$&$2$ & $0$ & $0$ & $2$\\ \hline

Oulu-NPU~\cite{OULU_NPU_2017}&$2017$&$55$/$5,940$& Frontal& No &No&$1$&$1$& $0$  & $0$ & $0$ & $2$\\
 \hline

SiW~\cite{learning-deep-models-for-face-anti-spoofing-binary-or-auxiliary-supervision} &$2018$&$165$/$4,620$& $[-90^{\circ},90^{\circ}]$  & Yes &Yes&$1$&$1$& $0$ & $0$ & $0$ & $2$ \\ \hline

SiW-M &$2019$&$493$/$1,630$& $[-90^{\circ},90^{\circ}]$  & Yes & Yes & $1$ & $1$ & $5$ & $3$ & $3$ & $13$\\ \hline

\hline
	\end{tabular}
	} 

\label{tab:dataset}
\vspace{-5mm}
\end{table*}

\Paragraph{Face Anti-spoofing}
Image-based face anti-spoofing refers to face anti-spoofing techniques that only take RGB images as input without extra information such as depth or heat. 
In early years, researchers utilize liveness cues, such as eye blinking and head motion, to detect print attacks~\cite{pan2007eyeblink,patel2016cross,shao2017deep,kollreider2007real}. 
However, when encountering unknown attacks, such as photograh with eye portion cut, and video replay, those methods suffer from a total failure.
Later, research move to a more general texture analysis and address print and replay attacks. 
Researchers mainly utilize handcrafted features, e.g., LBP~\cite{de2012lbp,de2013can,maatta2011face,boulkenafet2015face}, HoG~\cite{komulainen2013context,yang2013face}, SIFT~\cite{patel2016secure} and SURF~\cite{boulkenafet2017face}, with traditional classifiers, e.g., SVM and LDA, to make a binary decision. 
Those methods perform well on the testing data from the same database.
However, while changing the testing conditions such as lighting and background, they often have a large performance drop, which can be viewed as an overfitting issue.
Moreover, they also show limitations in handling $3$D mask attacks, mentioned in~\cite{liu20163d}. 

To overcome the overfitting issue, researchers make various attempts. 
Boulkenafet et al.~extract the spoofing features in HSV$+$YCbCR space~\cite{boulkenafet2015face}. 
Works in~\cite{agarwal2016face,bao2009liveness,bharadwaj2014face,feng2016integration,xu2015learning} consider features in the temporal domain.
Recent works~\cite{face-anti-spoofing-using-patch-and-depth-based-cnns,agarwal2016face} augment the data by using image patches, and fuse the scores from patches to a single decision.
For $3$D mask attacks, the heart pulse rate is estimated to differentiate $3$D mask from real faces~\cite{li2016generalized, liu20163d}. 
In the deep learning era, researchers propose several CNN works~\cite{face-anti-spoofing-using-patch-and-depth-based-cnns,Jourabloo_2018_ECCV,learning-deep-models-for-face-anti-spoofing-binary-or-auxiliary-supervision, feng2016integration,li2016original, patel2016cross, yang2014learn} that outperform the traditional methods.

\Paragraph{Zero-shot learning and unknown spoof attacks}
Zero-shot object recognition, or more generally, zero-shot learning, aims to recognize objects from unknown classes~\cite{socher2013zero}, i.e., object classes unseen in training. 
The overall idea is to associate the known and unknown classes via a semantic embedding, whose embedding spaces can be attributes~\cite{lampert2009learning}, word vector~\cite{frome2013devise}, text description~\cite{zhang2017learning} and human gaze~\cite{karessli2017gaze}. 

Zero-shot learning for unknown spoof attack, i.e., ZSFA, is a relatively new topic with unique properties. 
Firstly, unlike zero-shot object recognition, ZSFA emphasizes the detection of spoof attacks, instead of recognizing specific spoof types. 
Secondly, unlike generic objects with rich semantic embedding, there is no explicit well-defined semantic embedding for spoof patterns~\cite{Jourabloo_2018_ECCV}. 
As elaborated in Sec.~\ref{sec:intro}, prior ZSFA works~\cite{face-upad,arashloo2017anomaly} only model the live data via handcrafted features and standard generative models, with several drawbacks.
In this work, we propose a deep tree network to unsupervisely learn the semantic embedding for known spoof attacks. 
The partition of the data naturally associates certain semantic attributes with the sub-groups. 
During the testing, the unknown attacks are projected to the embedding to find the closest attributes for spoof detection.

\Paragraph{Deep tree networks}
Tree structure is often found helpful in tackling language-related tasks such as parsing and translation~\cite{chen2018tree}, due to the intrinsic relation of words and sentences. 
E.g., tree models are applied to joint vision and language problems such as visual question reasoning~\cite{cao2018visual}. 
Tree structure also has the property for learning features hierarchically. 
Face alignment works~\cite{kazemi2014one,valle2018deeply} utilize the regression trees to estimate facial landmarks from coarse to fine.
Xiong et al.~propose a tree CNN to handle the large-pose face recognition~\cite{xiong2015conditional}. 
In~\cite{kaneko2018generative}, Kaneko et al.~propose a GAN with decision trees to learn hierarchically interpretable representations. 
In our work, we utilize tree networks to learn the latent semantic embedding for ZSFA.

\Paragraph{Face anti-spoofing databases}
Given the significance of a good-quality database, researchers have released several face anti-spoofing databases, such as CASIA-FASD~\cite{zhang2012face}, Replay-Attack~\cite{Chingovska_BIOSIG_2012}, OULU-NPU~\cite{OULU_NPU_2017}, and SiW~\cite{learning-deep-models-for-face-anti-spoofing-binary-or-auxiliary-supervision} for print/replay attacks, and HKBU-MARs~\cite{liu20163d} for $3$D mask attacks. 
Early databases such as CASIA-FASD and Replay-Attack~\cite{zhang2012face} have limited subject variety, pose/expression/lighting variations, and video resolutions. 
Recent databases~\cite{OULU_NPU_2017,liu20163d,learning-deep-models-for-face-anti-spoofing-binary-or-auxiliary-supervision} improve those aspects, and also set up diverse evaluation protocols. 
However, up to now, all databases focus on either print/replay attacks, or $3$D mask attacks. 
To provide a comprehensive study of face anti-spoofing, especially the challenging ZSFA, we for the first time collect the database with diverse types of spoof attacks, as in Tab.~\ref{tab:dataset}. 
The details of our database are in Sec.~\ref{sec:database}.

\begin{figure*}[t!]
\centering
\includegraphics[width=0.9\linewidth]{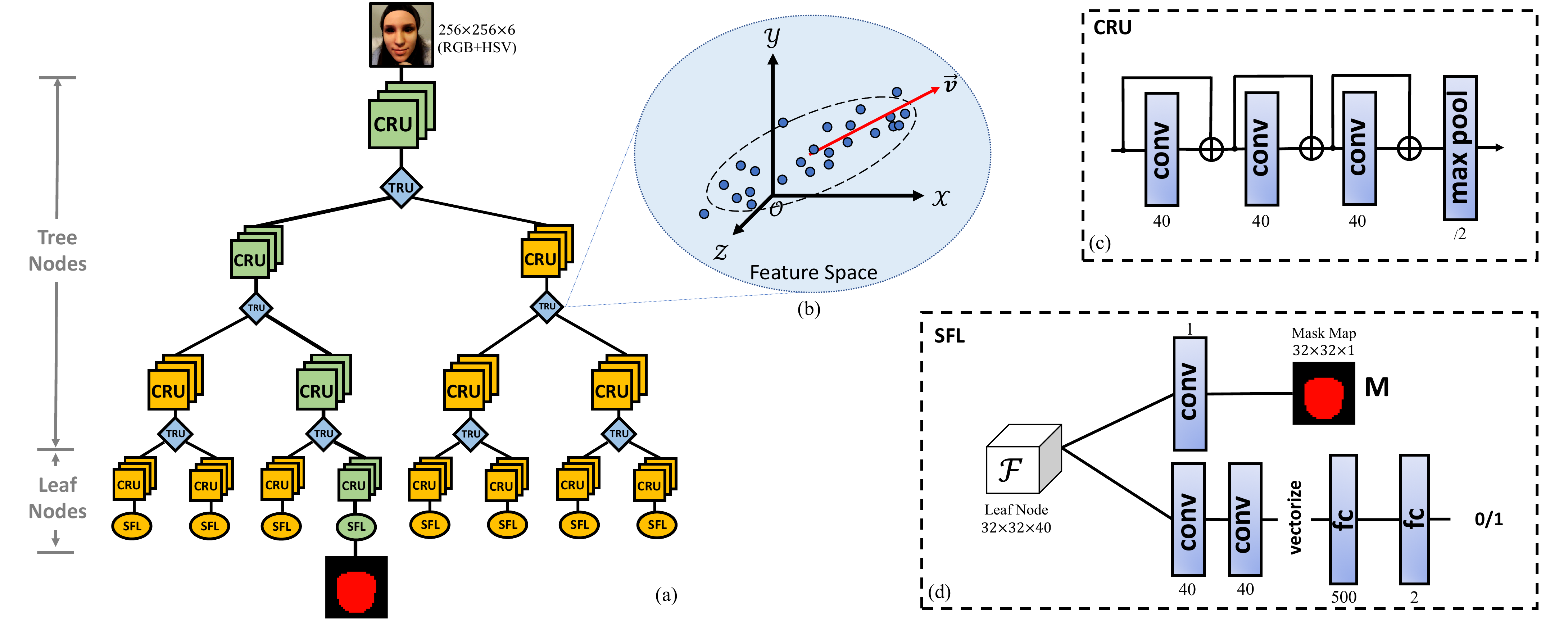}
\vspace{-2mm}
\caption{\small The proposed Deep Tree Network (DTN) architecture. (a) the overall structure of DTN. A tree node consists of a Convolutional Residual Unit (CRU) and a Tree Routing Unit (TRU), and a leaf node consists of a CRU and a Supervised Feature Learning (SFL) module. (b) the concept of Tree Routing Unit (TRU): finding the base with largest variations; (c) the structure of each Convolutional Residual Unit (CRU); (d) the structure of the Supervised Feature Learning (SFL) in the leaf nodes.  }
\label{fig:arch}
\figvspace 
\end{figure*}

\Section{Deep Tree Network for ZSFA}
\label{sec:alg}
The main purposes of DTN are twofold: $1)$ discover the semantic sub-groups for known spoofs; $2)$ learn the features in a hierarchical way. 
The architecture of DTN is shown in Fig.~\ref{fig:arch}.
Each tree node consists of a Convolutional Residual Unit (CRU) and a Tree Routing Unit (TRU), while the leaf node consists of a CRU and a Supervised Feature Learning (SFL) module.
CRU is a block with convolutional layers and the short-cut connection.
TRU defines a node routing function to route a data sample to one of the child nodes. 
The routing function partitions all visiting data along the direction with the largest data variation.
SFL module concatenates the classification supervision and the pixel-wise supervision to learn the spoofing features.


\SubSection{Unsupervised Tree Learning}
\label{sec:tree}
\vspace{+2mm}
\SubSubSection{Node Routing Function}

For a TRU node, let's assume the input $\textbf{\textit{x}} = f(\mathbf{I} \, | \, \theta) \in \mathbb{R}^{m}$ is the vectorized feature response, $\mathbf{I}$ is data input, $\theta$ is the parameters of the previous CRUs, and $\mathcal{S}$ is the set of data samples $\mathbf{I}_k, k = 1,2,...,K$ that visit this TRU node.
In~\cite{xiong2015conditional}, Xiong et al.~define a routing function as:
\begin{equation}\label{eq:old_split}
\varphi(\textbf{\textit{x}}) = \textbf{\textit{x}}^T \cdot \textbf{\textit{v}} + \tau,
\eqnvspace
\end{equation}
where $\textbf{\textit{v}}$ denotes the projection vector and $\tau$ is the bias. Data $\mathcal{S}$  can then be split into $\mathcal{S}_{left}:\{\mathbf{I}_k | \varphi(\textbf{\textit{x}}_k) < 0 , \mathbf{I}_k \in \mathcal{S}\}$ and $\mathcal{S}_{right}:\{\mathbf{I}_k | \varphi(\textbf{\textit{x}}_k) \geq 0 , \mathbf{I}_k \in \mathcal{S} \}$, and directed to the left and right child node, respectively. 
To learn this function, they propose to maximize the distance between the mean of $\mathcal{S}_{left}$ and $\mathcal{S}_{right}$, while keeping the mean of $\mathcal{S}$ centered at $0$. 
This unsupervised loss is formulated as:
\begin{equation}\label{eq:old_loss}
\mathcal{L} = \frac{
(\frac{1}{N}\sum\limits_{I_k\in \mathcal{S}}
\varphi(\textbf{\textit{x}}_k))^2}
{(\frac{1}{N_l}\sum\limits_{I_k\in \mathcal{S}_{left}} \varphi(\textbf{\textit{x}}_k)-\frac{1}{N_r}\sum\limits_{I_k\in \mathcal{S}_{right}} \varphi(\textbf{\textit{x}}_k))^2},
\eqnvspace
\end{equation} 
where $N$, $N_l$, $N_r$ denote the number of samples in each set.  

However, in practice, minizing Equ.~\ref{eq:old_loss} might not lead to a satisfactory solution. 
Firstly, the loss can be minimized by increasing the norm of either $\textbf{\textit{v}}$ or $\textbf{\textit{x}}$, which is a trivial solution. 
Secondly, even when the norms of {$\textbf{\textit{v}}$, $\textbf{\textit{x}}$} are constrained, Equ.~\ref{eq:old_loss} is affected by the density of data $\mathcal{S}$ and can be sensitive to the outliers.
In other words, the zero expectation of $\varphi(\textbf{\textit{x}})$ does not necessarily result in a balanced partition of data $\mathcal{S}$.
Local minima could be achieved when all data are split to one side.
In some cases, the tree may suffer from collapsing to a few (even one) leaf nodes.

To better partition the data, we propose a novel routing function and an unsupervised loss. 
Regardless of $\tau$, the dot product between $\textbf{\textit{x}}^T$ and $\textbf{\textit{v}}$ can be regarded as projecting $\textbf{\textit{x}}$ to the direction of $\textbf{\textit{v}}$.
We design $\textbf{\textit{v}}$ such that we can observe the largest variation after projection.
Inspired by the concept of PCA, the optimal solution naturally becomes the largest PCA basis of data $\mathcal{S}$. 
To achieve this, we first constrain $\textbf{\textit{v}}$ to be norm $1$ and reformulate Equ.~\ref{eq:old_split} as:
\begin{equation}\label{eq:new_split}
\varphi(\textbf{\textit{x}}) = (\textbf{\textit{x}}-\boldsymbol{\mu})^T \cdot \textbf{\textit{v}},  \quad 
\norm{\textbf{\textit{v}}}=1,
\end{equation} 
where $\boldsymbol{\mu}$ is the mean of data $\mathcal{S}$.  
Then, finding $\textbf{\textit{v}}$ is identical to finding the largest eigenvector of the covariance matrix $\bar{\textbf{\textit{X}}}^T_\mathcal{S} \bar{\textbf{\textit{X}}}_\mathcal{S}$, where $\bar{\textbf{\textit{X}}}_\mathcal{S} = \textbf{\textit{X}}_\mathcal{S} - \boldsymbol{\mu}$, and $\textbf{\textit{X}}_\mathcal{S} \in \mathbb{R}^{N \times K}$ is the data matrix. Based on the definition of eigen-analysis $\bar{\textbf{\textit{X}}}^T_\mathcal{S} \bar{\textbf{\textit{X}}}_\mathcal{S} \textbf{\textit{v}} = \lambda \textbf{\textit{v}}$, our optimization aims to maximize:
\begin{gather}\label{eq:new_opt}
\argmax_{\textbf{\textit{v}},\theta} \, \lambda 
= \argmax_{\textbf{\textit{v}},\theta} \, \textbf{\textit{v}}^T\bar{\textbf{\textit{X}}}^T_\mathcal{S} \bar{\textbf{\textit{X}}}_\mathcal{S}\textbf{\textit{v}}.
\end{gather} 
The loss for learning the routing function is formulated as:
 \begin{gather}\label{eq:new_loss}
\mathcal{L}_{route} = \textrm{exp}(-\alpha\textbf{\textit{v}}^T\bar{\textbf{\textit{X}}}^T_\mathcal{S} \bar{\textbf{\textit{X}}}_\mathcal{S}\textbf{\textit{v}}) + \beta \textrm{Tr}(\bar{\textbf{\textit{X}}}^T_\mathcal{S} \bar{\textbf{\textit{X}}}_\mathcal{S}),
\end{gather} 
where $\alpha,\beta$ are scalars, and set as $1$e-$3$, $1$e-$2$ in our experiments. 
We apply the exponential function on the first term to make the maximization problem bounded. The second term is introduced as a regularizer to prevent trivial solutions by constraining the trace of covariance matrix of $\bar{\textbf{\textit{X}}}_\mathcal{S}$.

\SubSubSection{Tree of Known Spoofs}
With the routing function, we can build the entire binary tree. 
Fig.~\ref{fig:arch} shows a binary tree of depth of $4$, with $8$ leaf nodes.
As mentioned early in Sec.~\ref{sec:alg}, the tree is designed to find the semantic sub-groups from all known spoofs, and is termed as spoof tree.
Similarly, we may also train live tree with live faces only, as well as general data tree with both live and spoof data.
Compared to spoof tree, live and general data tree have some drawbacks.
Live tree does not convey semantic meaning for the spoof, and the attributes learned at each node cannot help to route and better detect spoof; 
General data tree may result in imbalanced sub-groups, where samples of one class outnumber another. 
Such imbalance would cause bias for supervised learning in the next stage.

Hence, when we compute Equ.~\ref{eq:new_loss} to learn the routing functions, we only consider the spoof samples to construct $\textbf{\textit{X}}_\mathcal{S}$.
To have a balanced sub-group for each leaf, we suppress the responses of live data to zero, so that all live data can be evenly partitioned to the child nodes.
Meanwhile, we also suppress the responses of the spoof data that do not visit this node, so that every node models the distribution of a unique spoof subset.

Formally, for each node, we maximize the routing function responses of spoof data that visit this node (denoted as $\mathcal{S}$), while minimizing the responses of other data (denoted as $\mathcal{S}^-$), including all live data and spoof data that don't visit this node, i.e., that visit neighboring nodes. 
To achieve this objective, we define the following loss: 
\begin{gather}\label{eq:uniq_loss}
\mathcal{L}_{uniq} = 
-\frac{1}{N}\sum\limits_{\mathbf{I}_k \in \mathcal{S}}
\left\|\bar{\textbf{\textit{x}}}^T_k \textbf{\textit{v}} \right\|^2 + 
\frac{1}{N^-}\sum\limits_{\mathbf{I}_k \in \mathcal{S}^-}
\left\| \bar{\textbf{\textit{x}}}^T_k \textbf{\textit{v}} \right\|^2.
\end{gather} 

\SubSection{Supervised Feature Learning}
Given the routing functions, a data sample $\mathbf{I}_k$ will be assigned to one of the leaf nodes. 
Let's first define the feature output of leaf node as $\mathcal{F}(\mathbf{I}_k \, | \, \theta)$, shortened as $\mathcal{F}_k$ for simplicity.
At each leaf node, we define two node-wise supervised tasks to learn discriminative features: $1)$ binary classification drives the learning of a high-level understanding of live vs.~spoof faces, 
$2)$ pixel-wise mask regression draws CNN's attention to low-level local feature learning.

\Paragraph{Classification supervision}
\label{sec:super}
To learn a binary classifier, as shown in Fig.~\ref{fig:arch}(d), we apply two additional convolution layers and two fully connected layers on $\mathcal{F}_k$ to generate  a feature vector $\mathbf{c}_k \in \mathbb{R}^{500}$.  
We supervise the learning via the softmax cross entropy loss:
\begin{gather} \label{eq:soft_loss}
\mathcal{L}_{class} = \frac{1}{N}
\sum\limits_{I_k \in \mathcal{S}}
{\Big\{
(1-y_k)      \textrm{log}      (1-p_k ) -y_k      \textrm{log}      p_k    
\Big\}} \\
p_k =  \frac{ \textrm{exp}({\mathbf{w}_1}^T \mathbf{c}_k) }   {\textrm{exp}({\mathbf{w}_0}^T \mathbf{c}_k)+\textrm{exp}({\mathbf{w}_1}^T \mathbf{c}_k)},
\end{gather}
where $\mathcal{S}$ represents all the data samples that arrive this leaf node, 
$N$ denotes the number of samples in $\mathcal{S}$, $\{\mathbf{w}_0,\mathbf{w}_1\}$ are the parameters in the last fully connected layer, and $y_k$ is the label of data sample $k$ ($1$ denotes spoof, and $0$ live).

\begin{figure*}[t!]
\centering
\includegraphics[width=0.99\linewidth]{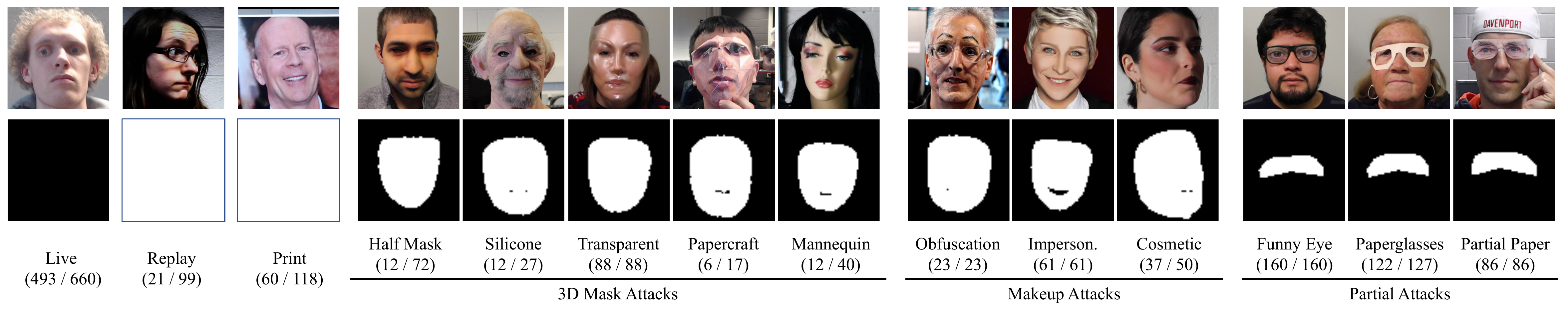}
\vspace{-3mm}
\caption{\small The examples of the live faces and $13$ types of spoof attacks. The second row shows the ground truth masks for the pixel-wise supervision $\textbf{D}_k$. For $(m,n)$ in the third row, $m/n$ denotes the number of subjects/videos for each type of data.}
\label{fig:label}
\figvspace 
\end{figure*}

\Paragraph{Pixel-wise supervision}
We also concatenate another convolution layer to $\mathcal{F}_k$ to generate a map response $\mathbf{M}_k \in \mathbb{R}^{32 \times 32}$.
Inspired by the prior work~\cite{learning-deep-models-for-face-anti-spoofing-binary-or-auxiliary-supervision}, we leverage the semantic prior knowledge of face shapes and spoof attack position to provide a pixel-wise supervision. 
Using the dense face alignment model~\cite{liu2017dense}, we provide a binary mask $\textbf{D}_k \in \mathbb{R}^{32 \times 32}$, shown in Fig.~\ref{fig:label}, to indicate the pixels of spoof mediums.  
Thus, for a leaf node, the loss function for the pixel-wise supervision is:
\begin{gather}\label{eq:mask_loss}
\mathcal{L}_{mask} = \frac{1}{N}\sum\limits_{I_k \in \mathcal{S}}
\left\|\mathbf{M}_k - \textbf{D}_k \right\|_1.
\end{gather} 

\Paragraph{Overall loss}
Finally, we apply the supervised losses on $p$ leaf nodes, the unsupervised losses on $q$ TRU nodes, and formulate our training loss as:
\begin{equation}\label{eq:overall_loss}
\mathcal{L} =  \sum\limits_{i = 1}^{p}  (  \alpha_1\mathcal{L}_{class}^i  + 
\alpha_2   \mathcal{L}_{mask}^i  )+ 
\sum\limits_{j = 1}^{q} ( \alpha_3 \mathcal{L}_{route}^j + \alpha_4 \mathcal{L}_{uniq}^j),
\end{equation} 
where $\alpha_1$,$\alpha_2$,$\alpha_3$,$\alpha_4$ are the regularization coefficients for each term, and are set as $0.001$, $1.0$, $2.0$, $0.001$ respectively. For a $4$-layer DTN, $p=8$ and $q=7$. 

\begin{figure}[t!]
\centering
\includegraphics[width=0.9\linewidth]{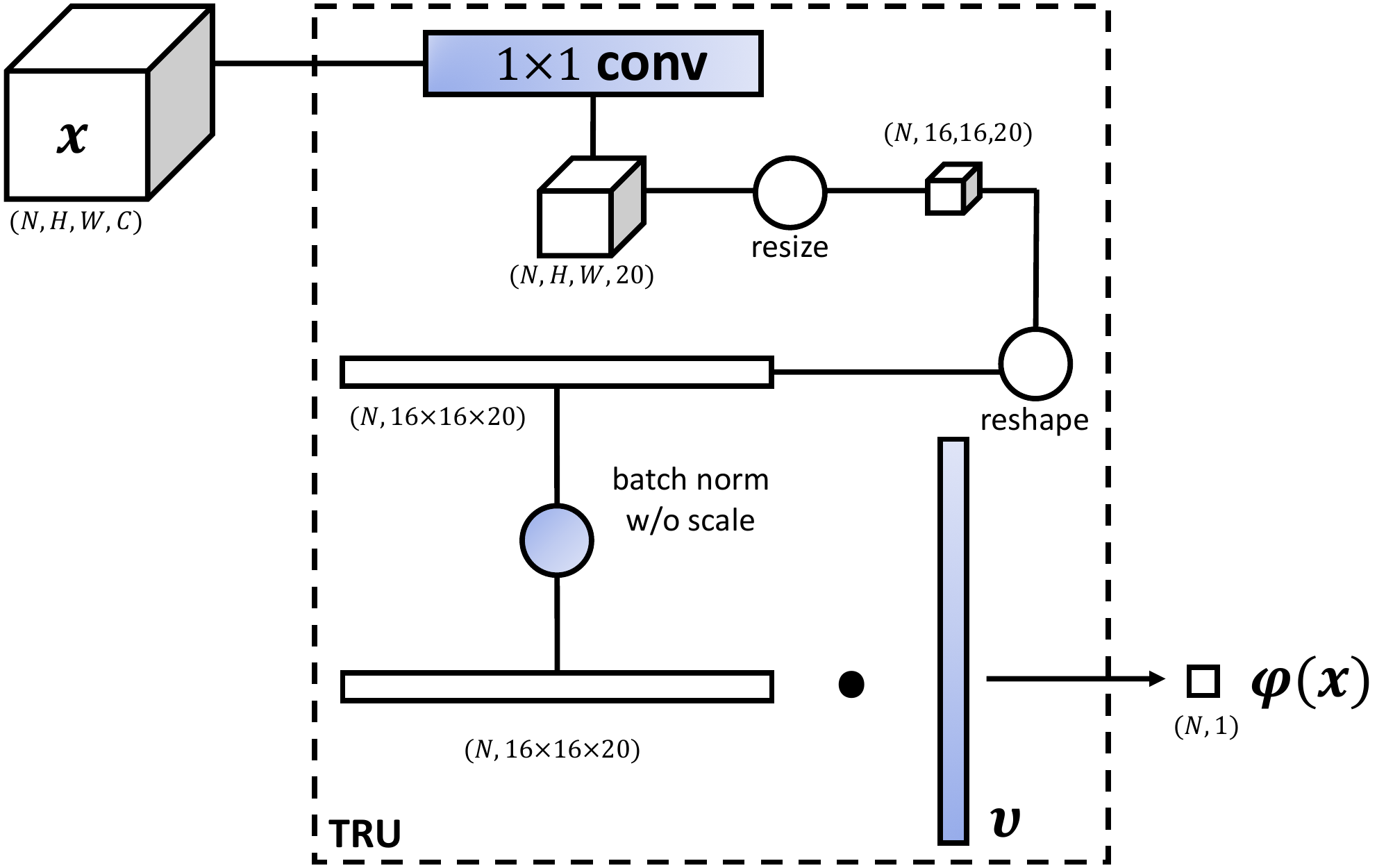}
\caption{\small The structure of the Tree Routing Unit (TRU). }
\label{fig:tru}
\vspace{-4mm}
\end{figure}

\SubSection{Network Architecture}
\label{sec:net}


\Paragraph{Deep Tree Network (DTN)}
DTN is the main framework of the proposed model. 
It takes $\mathbf{I} \in \mathbb{R}^{256 \times 256 \times 6}$ as input, where the $6$ channels are RGB+HSV color spaces.
We concatenate three $3\times 3$ convolution layers with $40$ channels and $1$ max-pooling layer, and group them as one Convolutional Residual Unit (CRU). 
Each convolution layer is equipped with ReLU and group normalization layer~\cite{Wu_2018_ECCV}, due to the dynamic batch size in the network. 
We also apply a shortcut connection for each convolution layer.
For each tree node, we deploy one CRU before the TRU. 
At the leaf node, DTN produces the feature representation of input $\mathbf{I}$ as $\mathcal{F}(\mathbf{I} \, | \, \theta) \in \mathbb{R}^{32 \times 32 \times 40}$, then uses one $1 \times 1$ convolution layer to generate the binary mask map $\mathbf{M}$.

\Paragraph{Tree Routing Unit (TRU)}
TRU is the module routing the data sample to one of the child CRUs. 
As shown in Fig.~\ref{fig:tru}, it first compresses the feature by using an $1 \times 1$ convolution layer, and resizing the response spatially. 
For the root node, we compress the CRU feature to $\textbf{x}\in \mathbb{R}^{32 \times 32 \times 10}$, and for later tree node, we compress the CRU feature to $\textbf{x} \in \mathbb{R}^{16 \times 16 \times 20}$.
Compressing the input feature to a smaller size  helps to reduce the burden of computating and saving the covariance matrix in Equ.~\ref{eq:new_loss}. 
E.g., the vectorized feature for the first CRU is $\textbf{\textit{x}} \in \mathbb{R}^{655,360}$, and the covariance matrix of $\textbf{\textit{x}}$ can take $\sim 400$GB in memory. 
However, after compression the vectorized feature is $\textbf{\textit{x} }\in \mathbb{R}^{10,240}$, and the covariance matrix of $\textbf{\textit{x}}$ only needs $\sim 0.1$GB of memory.

After that, we vectorize the output and apply the routing function $\varphi(\textbf{\textit{x}})$.
To compute $\boldsymbol{\mu}$ in Equ.~\ref{eq:new_split}, instead of optimizing it as a variable of the network, we simply apply a batch normalization layer without scaling to save the moving average of each mini-batch.
In the end, we project the compressed CRU response to the largest basis $\textbf{\textit{v}}$ and obtain the projection coefficient. 
Then we assign the samples with negative coefficient to the left child CRU and the samples with  positive coefficient to the right child CRU.

\Paragraph{Implementation details}
With the overall loss in Equ.~\ref{eq:overall_loss}, our proposed network is trained in an end-to-end fashion. 
All losses are computed based on each mini-batch.
DTN modules and TRU modules are optimized alternately. 
While optimizing DTN, we keep the parameters of TRUs fixed and vice versa.

\Section{Spoof in the Wild Database with Multiple Attack Types}
\label{sec:database}
To benchmark face anti-spoofing methods specifically for unknown attacks, we collect the Spoof in the Wild database with Multiple Attack Types (SiW-M). 
Compared with the previous databases in Tab.~\ref{tab:dataset}, SiW-M shows a great diversity in spoof attacks, subject identities,  environments and other factors. 

For spoof data collection, we consider two spoofing scenarios: \textit{impersonation}, which entails the use of spoof to be recognized as someone else, and \textit{obfuscation}, which entails the use to remove the attacker's own identity. 
In total, we collect $968$ videos of $13$ types of spoof attacks listed hieratically in Fig~\ref{fig:label}. 
For all $5$ mask attacks, $3$ partial attacks, obfuscation makeup and cosmetic makeup, we record $1080$P HD videos.
For impersonation makeup, we collect $720$P videos from Youtube due to the lack of special makeup artists.
For print and replay attacks, we intend to collect videos from harder cases where the existing system fails. 
Hence, we deploy an off-the-shelf face anti-spoofing algorithm~\cite{learning-deep-models-for-face-anti-spoofing-binary-or-auxiliary-supervision} and record spoof videos when the algorithm predicts live.

For live data, we include $660$ videos from $493$ subjects. 
In comparison, the number of subjects in SiW-M is $9$ times larger than Oulu-NPU~\cite{OULU_NPU_2017} and CASIA-FASD~\cite{zhang2012face}, and $3$ times larger than SiW~\cite{learning-deep-models-for-face-anti-spoofing-binary-or-auxiliary-supervision}.
In addition, subjects are diverse in ethnicity and age.
The live videos are collected in $3$ sessions: $1)$ a room environment where the subjects are recorded with few variations such as pose, lighting and expression (PIE). $2)$ a different and much larger room where the subjects are also recorded with PIE variations. $3)$ a mobile phone mode, where the subjects are moving while the phone camera is recording. Extreme pose angles and lighting conditions are introduced. Similar to print and replay videos, we deploy the face anti-spoofing algorithm~\cite{learning-deep-models-for-face-anti-spoofing-binary-or-auxiliary-supervision} to find out the videos where the algorithm predicts spoof. 
Hence, this third session is a harder scenario. 

In total, we collect $1,630$ videos and each lasts $5$-$7$ seconds. 
The $1080$P videos are recorded by Logitech C$920$ webcam and Canon EOS T$6$.
To use SiW-M for the study of ZSFA, we define the leave-one-out testing protocols.
Each time we train a model with $12$ types of spoof attacks plus the $80\%$ of the live videos, and test on the left $1$ attack type plus the $20\%$ of live videos.
There is no overlapping subjects between the training and testing sets of live videos.

\begin{table*}[t!]
\small
	\centering
	\caption{ AUC ($\%$) of the model testing on CASIA, Replay, and MSU-MFSD.}\vspace{-3mm}
	\resizebox{\textwidth}{!} 
{
	\begin{tabular}{l|c|c|c|c|c|c|c|c|c|c}
	\hline
	\multirow{2}{*}{Methods} & \multicolumn{3}{c|}{CASIA~\cite{zhang2012face}} & \multicolumn{3}{c|}{Replay-Attack~\cite{Chingovska_BIOSIG_2012}} &  \multicolumn{3}{c|}{MSU~\cite{wen2015face}}  & \multirow{2}{*}{Overall}\\ \cline{2-10}
	 &Video&Cut Photo & Warped Photo & Video & Digital Photo & Printed Photo & Printed Photo & HR Video & Mobile Video &\\ \hline	\hline
   OC-SVM$_{RBF}$+BSIF~\cite{arashloo2017anomaly} &$70.7$& $60.7$ & $95.9$ & $84.3$ & $88.1$ & $73.7$ & $64.8$ & $87.4$ & $74.7$ & $78.7\pm11.7$   \\ \cline{1-11}
   SVM$_{RBF}$+LBP~\cite{OULU_NPU_2017} &$91.5$& $91.7$ & $84.5$ & $99.1$ & $98.2$ & $87.3$ & $47.7$ & $99.5$ & $\mathbf{97.6}$ & $88.6\pm16.3$  \\ \cline{1-11}
   NN+LBP~\cite{face-upad} &$\mathbf{94.2}$& $88.4$ & $79.9$ & $99.8$ & $95.2$ & $78.9$ & $50.6$ & $99.9$ & $93.5$ & $86.7\pm15.6$   \\ \cline{1-11}  \hline\hline
      Ours &$90.0$& $\mathbf{97.3}$ & $\mathbf{97.5}$ & $\mathbf{99.9}$ & $\mathbf{99.9}$ & $\mathbf{99.6}$ & $\mathbf{81.6}$ & $\mathbf{99.9}$ & $97.5$ & $\mathbf{95.9}\pm\mathbf{6.2}$  \\ \cline{1-11}
    \hline\hline
	\end{tabular}
	} 
\label{tab:intra}
\figvspace
\end{table*}

\Section{Experimental Results} \label{sec:exp}

\SubSection{Experimental Setup}
\paragraph{Databases}
We evaluate our proposed method on multiple databases. We deploy the leave-one-out testing protocols on SiW-M and report the results of $13$ experiments. 
Also, we test on previous face anti-spoofing databases, including CASIA~\cite{zhang2012face}, Replay-Attack~\cite{Chingovska_BIOSIG_2012}, and MSU-MFSD~\cite{wen2015face}), compare with the state of the art.

\paragraph{Evaluation metrics}
\vspace{-5mm}
We evaluate with the following metrics: Attack Presentation Classification Error Rate (APCER)~\cite{acer1}, Bona Fide Presentation Classification Error Rate (BPCER)~\cite{acer1}, the average of APCER and BPCER, Average Classification Error Rate (ACER)~\cite{acer1}, Equal Error Rate (EER), and Area Under Curve (AUC). 
Note that, in the evaluation of unknown attacks, we assume there is no validation set to tune the model and thresholds while calculating the metrics. 
Hence, we determine the threshold based on the training set and fix it for all testing protocols.
A single test sample is one video frame, instead of one video.

\begin{table}[t!]
\small
	\caption{Compare models with different routing strategies.}\vspace{-3mm}
	\resizebox{0.5\textwidth}{!} 
{
	\begin{tabular}{c|c|c|c|c}
	\hline
	Strategies & APCER & BPCER & ACER& EER\\ \hline \hline
	Random routing & $37.1$ & $\mathbf{16.1}$ & $26.6$ & $24.7$ \\ \hline
	Pick-one-leaf & $51.2\pm20.0$ & $18.1 \pm4.9$& $34.7\pm8.8$ & $24.1\pm 3.1$ \\ \hline
	Proposed routing function & $\mathbf{17.0}$ & $21.5$ & $\mathbf{19.3}$ & $\mathbf{19.8}$ \\ \hline
    \hline
	\end{tabular}
     } 
\label{tab:ab1}
\end{table}

\begin{table}[t!]
\small
	\caption{Compare models with different tree losses and strategies. The first two terms of row $2$-$5$ refer to using live or spoof data in tree learning.  The last row is our method.} \vspace{-3mm}
	\resizebox{0.5\textwidth}{!} 
{
	\begin{tabular}{c|c|c|c|c}
	\hline
	Methods & APCER & BPCER & ACER& EER\\ \hline \hline
	MPT~\cite{xiong2015conditional} & $31.4$ & $24.2$ & $27.8$ & $27.3$ \\ 
	Live data $\surd$, Spoof data $\surd$, Unique Loss $\times$ & $~\mathbf{1.4}$ & $73.3$ & $37.3$ & $31.2$ \\ 
	Live data $\times$, Spoof data $\surd$, Unique Loss $\times$ & $70.0$ & $12.7$ & $41.3$ & $44.8$ \\ 
	Live data $\surd$, Spoof data $\surd$, Unique Loss $\surd$ & $54.2$ & $~\mathbf{12.5}$ & $33.4$ & $36.2$ \\ \hline
	Live data $\times$, Spoof data $\surd$, Unique Loss $\surd$ & $17.0$ & $21.5$ & $~\mathbf{19.3}$ & $~\mathbf{19.8}$ \\ \hline
    \hline
	\end{tabular}
	} 
\label{tab:ab2}
\vspace{-5mm}
\end{table}

\paragraph{Parameter setting}
\vspace{-5mm}
The proposed method is implemented in Tensorflow, and trained with a constant learning rate of $0.001$ with a batch size of $32$.
It takes $15$ epochs to converge. 
We randomly initialize all the weights using a normal distribution of $0$ mean and $0.02$ standard deviation.

\SubSection{Experimental Comparison}
\vspace{+2mm}
\SubSubSection{Ablation Study}
\vspace{-1mm}
All ablation studies use the Funny Eye protocol.

\Paragraph{Different fusion methods}
In the proposed model, both the norm of the mask maps and binary spoof scores could be utilized for the final classification. 
To find the best fusion method, we compute ACER from using map norm, softmax score, the maximum of map norm and softmax score, and the average of two values, and obtain $31.7\%$, $20.5\%$, $21.0\%$, and $19.3\%$ respectively. 
Since the average score of the mask norm and binary spoof score performs the best,  we use it for the remaining experiments. 
Moreover,  we set $0.2$ as the final threshold to compute APCER, BPCER and ACER for all the experiments. 

\Paragraph{Different routing methods}
Routing is a crucial step to find the best subgroup to detect spoofness of a testing sample.
To show the effect of proper routing, we evaluate $2$ alternative routing strategies: random routing and pick-one-leaf. 
Random routing denotes randomly selecting one leaf node for a testing sample to produce prediction; 
Pick-one-leaf denotes constantly selecting one particular leaf node to produce results, for which we report the mean score and standard deviation of $8$ selections. 
Shown in Tab.~\ref{tab:ab1}, both strategies perform worse than the proposed routing function. 
In addition, the large standard deviation of pick-one-leaf strategy shows the {\it large} performance difference of $8$ subgroups on the {\it same type} of unknown attacks, and demonstrates the necessity of a proper routing.

\begin{table*}[t!]
\small
	\centering
	\caption{The evaluation and comparison of the testing on SiW-M.}
	\vspace{-3mm}
	\resizebox{\textwidth}{!} 
{
	\begin{tabular}{l|l|c|c|c|c|c|c|c|c|c|c|c|c|c|c}
	\hline
	\multirow{2}{*}{Methods} & \multirow{2}{*}{Metrics (\%)} &\multirow{2}{*}{Replay}& \multirow{2}{*}{Print} & \multicolumn{5}{c|}{Mask Attacks}  & \multicolumn{3}{c|}{Makeup Attacks}  &  \multicolumn{3}{c|}{Partial Attacks} & \multirow{2}{*}{Average}\\ \cline{5-15}
	 &&&  & Half & Silicone & Trans. & Paper & Manne. & Obfusc. & Imperson. & Cosmetic & Funny Eye & Paper Glasses & Partial Paper &\\ \hline	\hline
     \multirow{4}{*}{SVM$_{RBF}$+LBP~\cite{OULU_NPU_2017}} & APCER &$19.1$& $15.4$ & $40.8$ & $20.3$ & $70.3$ & $\mathbf{0.0}$ & $4.6$ & $96.9$ & $35.3$ & $\mathbf{11.3}$ & $53.3$ & $58.5$ & $0.6$ & $32.8\pm29.8 $  \\ \cline{3-16}
     &BPCER &$22.1$& $21.5$ & $21.9$ & $21.4$ & $20.7$ & $23.1$ & $22.9$ & $21.7$ & $12.5$ & $22.2$ & $18.4$ & $20.0$ & $22.9$ & $21.0\pm2.9$  \\ \cline{3-16} 
     &ACER &$20.6$& $18.4$ & $31.3$ & $21.4$ & $45.5$ & $11.6$ & $13.8$ & $59.3$ & $23.9$ & $16.7$ & $35.9$ & $39.2$ & $11.7$ & $26.9\pm14.5$  \\ \cline{3-16}
     &EER &$20.8$& $18.6$ & $36.3$ & $21.4$ & $37.2$ & $7.5$ & $14.1$ & $51.2$ & $19.8$ & $16.1$ & $34.4$ & $33.0$ & $7.9$ & $24.5\pm12.9$  \\ \hline	\hline
    \multirow{4}{*}{Auxiliary\cite{learning-deep-models-for-face-anti-spoofing-binary-or-auxiliary-supervision}} & APCER &$23.7$& $7.3$ & $27.7$ & $\mathbf{18.2}$ & $97.8$ & $8.3$ & $16.2$ & $100.0$ & $18.0$ & $16.3$ & $91.8$ & $72.2$ & $0.4$ & $38.3\pm37.4$  \\ \cline{3-16}
     & BPCER &$\mathbf{10.1}$& $\mathbf{6.5}$ & $\mathbf{10.9}$ & $\mathbf{11.6}$ & $\mathbf{6.2}$ & $\mathbf{7.8}$ & $\mathbf{9.3}$ & $11.6$ & $\mathbf{9.3}$ & $\mathbf{7.1}$ & $\mathbf{6.2}$ & $\mathbf{8.8}$ & $\mathbf{10.3}$ & $\mathbf{8.9}\pm\mathbf{2.0}$  \\ \cline{3-16}
     & ACER &$16.8$& $6.9$ & $19.3$ & $\mathbf{14.9}$ & $52.1$ & $8.0$ & $12.8$ & $55.8$ & $13.7$ & $\mathbf{11.7}$ & $49.0$ & $40.5$ & $\mathbf{5.3}$ &$23.6\pm18.5$  \\ \cline{3-16}
     & EER &$14.0$& $4.3$ & $\mathbf{11.6}$ & $\mathbf{12.4}$ & $\mathbf{24.6}$ & $7.8$ & $10.0$ & $72.3$ & $10.1$ & $\mathbf{9.4}$ & $21.4$ & $\mathbf{18.6}$ & $\mathbf{4.0}$  & $17.0\pm17.7$ \\ \hline	\hline
    \multirow{4}{*}{Ours} & APCER &$\textbf{1.0}$& $\mathbf{0.0}$ & $\mathbf{0.7}$ & $24.5$ & $\mathbf{58.6}$ & $0.5$ & $\mathbf{3.8}$ & $\mathbf{73.2}$ & $\mathbf{13.2}$ & $12.4$ & $\mathbf{17.0}$ & $\mathbf{17.0}$ & $\mathbf{0.2}$ & $\mathbf{17.1}\pm\mathbf{23.3}$  \\ \cline{3-16}
     & BPCER &$18.6$& $11.9$ & $29.3$ & $12.8$ & $13.4$ & $8.5$ & $23.0$ & $\mathbf{11.5}$ & $9.6$ & $16.0$ & $21.5$ & $22.6$ & $16.8$ & $16.6\pm6.2$  \\ \cline{3-16} 
     & ACER &$\mathbf{9.8}$& $\mathbf{6.0}$ & $\mathbf{15.0}$ & $18.7$ & $\mathbf{36.0}$ & $\mathbf{4.5}$ & $\mathbf{7.7}$ & $\mathbf{48.1}$ & $\mathbf{11.4}$ & $14.2$ & $\mathbf{19.3}$ & $\mathbf{19.8}$ & $8.5$ & $\mathbf{16.8}\pm\mathbf{11.1}$  \\ \cline{3-16}
     & EER &$\mathbf{10.0}$& $\mathbf{2.1}$ & $14.4$ & $18.6$ & $26.5$ & $\mathbf{5.7}$ & $\mathbf{9.6}$ & $\mathbf{50.2}$ & $\mathbf{10.1}$ & $13.2$ & $\mathbf{19.8}$ & $20.5$ & $8.8$ & $\mathbf{16.1}\pm\mathbf{12.2}$   \\ \cline{3-16}
    \hline\hline
	\end{tabular}
	} 
\label{tab:test}
\figvspace
\end{table*}

\Paragraph{Advantage of each loss function}
We have three important designs in our unsupervised tree learning: route loss $\mathcal{L}_{route}$, data used to compute the route loss, and the unique loss $\mathcal{L}_{uniq}$.
To show the effect of each loss and the training strategy, we train and compare networks with each loss excluded and alternative strategies. 
First, we train a network with the routing function proposed in~\cite{xiong2015conditional}, and then $4$ models with different modules on and off, shown in Tab.~\ref{tab:ab2}. 
The model with MPT~\cite{xiong2015conditional} routes data only to $2$ leaf nodes out of $8$ (i.e. tree collapse issue), which limits the performance.
Models without the unique loss exhibit the imbalance routing issue where sub-groups cannot be trained properly . 
Models using all data to learn the tree show worse performances than using spoof data only.
Finally, the proposed method performs the best among all options. 

\SubSubSection{Testing on existing databases}
\begin{figure}[t!]
\centering
\includegraphics[width=0.94\linewidth]{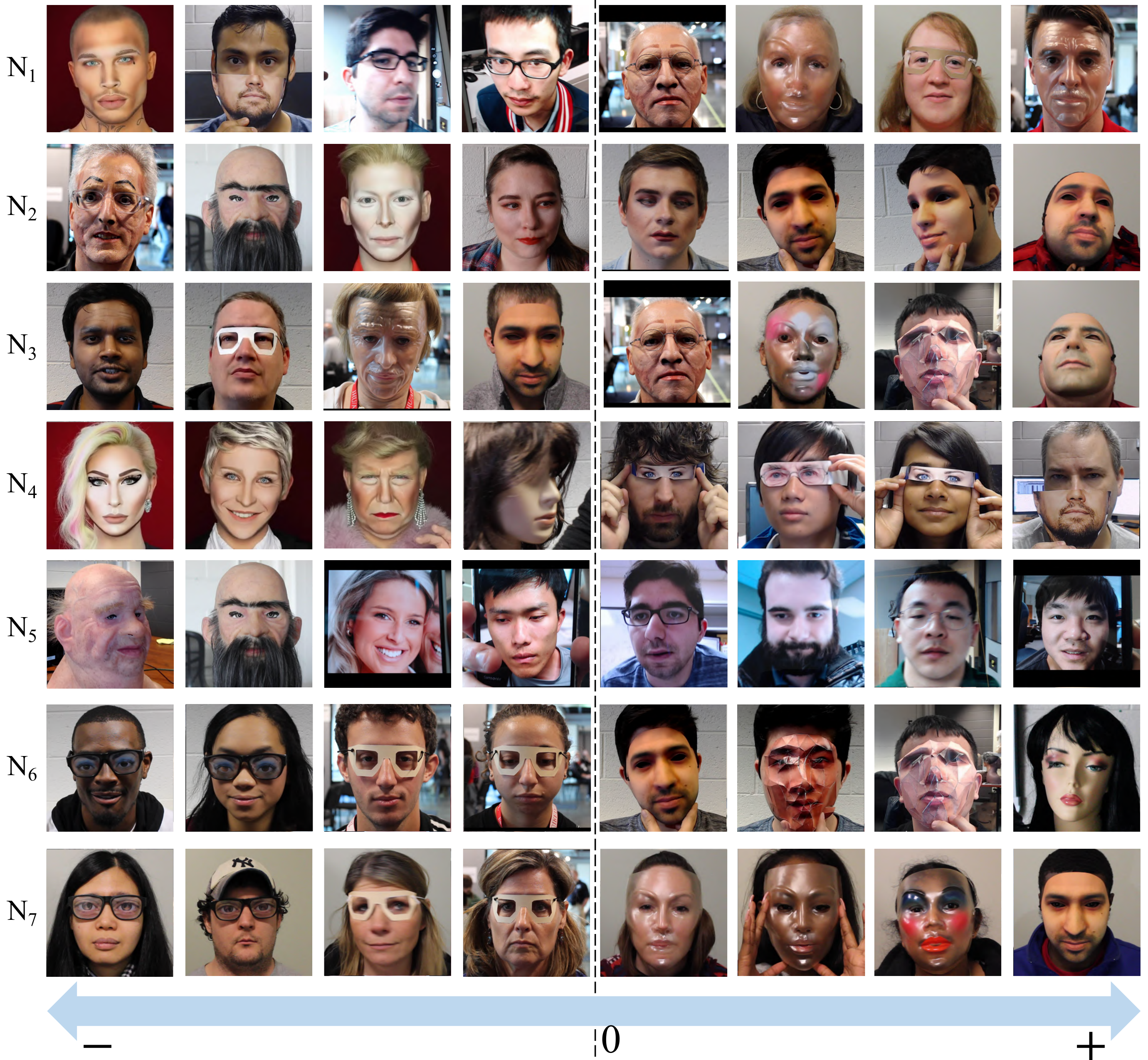}
\vspace{-3mm}
\caption{\small Visulization of the Tree Routing.}
\label{fig:vis1}
\figvspace 
\end{figure}

Following the protocol proposed in~\cite{arashloo2017anomaly}, we use CASIA~\cite{zhang2012face}, Replay-Attack~\cite{Chingovska_BIOSIG_2012} and MSU-MFSD~\cite{wen2015face} to perform ZSFA testing between replay and print attacks. 
Tab.~\ref{tab:intra} compares the proposed method with top three methods selected from over $20$ methods in~\cite{face-upad,arashloo2017anomaly,OULU_NPU_2017}. 
Our proposed method outperforms the prior state of the art by a convincing margin of $7.3\%$, and our smaller standard deviation further indicates a consistently good performance among unknown attacks. 

\SubSubSection{Testing on SiW-M}
We execute $13$ leave-one-out testing protocols on SiW-M. 
We compare with two of the most recent face anti-spoofing methods~\cite{OULU_NPU_2017,learning-deep-models-for-face-anti-spoofing-binary-or-auxiliary-supervision}, and set \cite{learning-deep-models-for-face-anti-spoofing-binary-or-auxiliary-supervision} as the baseline, which has demonstrated its SOTA performance on various benchmarks.
For a fair comparison with the baseline, we provide the same pixel-wise labeling (as in Fig.~\ref{fig:label}), and set the same threshold of $0.2$ to compute APCER, BPCER, and ACER. 

As shown in Tab.~\ref{tab:test}, our method achieves an overall better APCER, ACER and EER, with the improvement of baseline by $55\%$, $29\%$, and $5\%$. 
Specifically, we reduce the ACERs of transparent mask, funny eye, and paper glasses by $31\%$, $61\%$, and $51\%$, where the baseline models can be considered as total failures since they recognize most of the attacks as live.
Note that, ACER is more valuable in the context of ZSFA: 
no evaluation data for setting threshold and considerably varied thresholds for obtaining the EER performance. 
For instance, EERs of paper glasses model are similar between the baseline and our method, but with a preset threshold, our method offers a much better ACER.

Moreover, the proposed method is a more compact model than\cite{learning-deep-models-for-face-anti-spoofing-binary-or-auxiliary-supervision}. 
Given the input size of $256 \times 256 \times 6$, the baseline requires $87$ GFlops to compute the result while our method only needs $6$ GFlops  ($\times 15$ smaller). 
More analysis are shown with visualization in Sec.~\ref{sec:exp-vis}.

\begin{figure}[t!]
\centering
\includegraphics[width=0.88\linewidth]{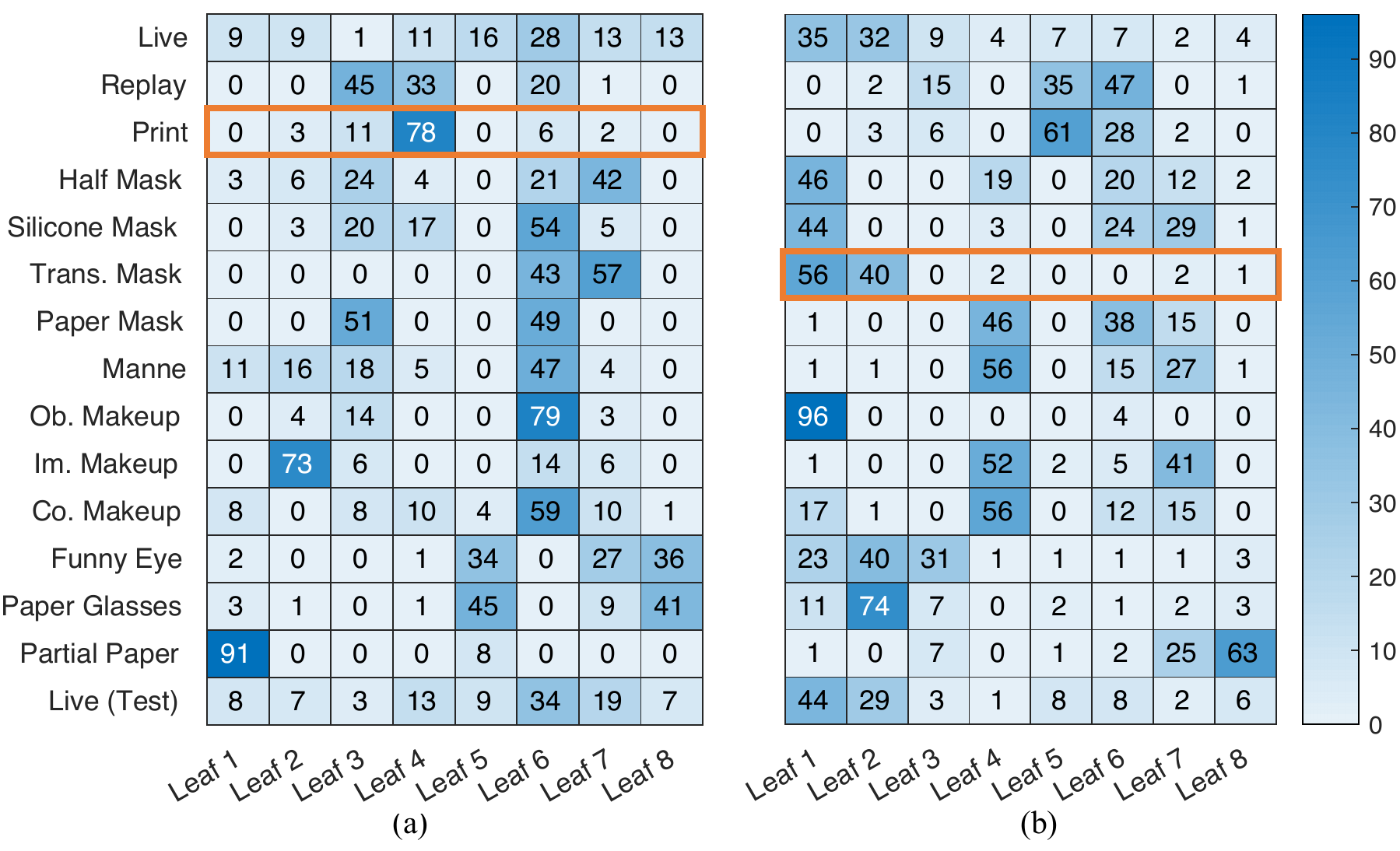}
\vspace{-3mm}
\caption{\small Tree routing distribution of live/spoof data. X-axis denotes $8$ leaf nodes, and y-axis denotes $15$ types of data. The number in each cell represents the percentage ($\%$) of data that fall in that leaf node. Each row is sum to $1$. (a) Print Protocol. (b) Transparent Mask Protocol. Yellow box denotes the unknown attacks.}
\label{fig:vis2}
\figvspace 
\end{figure}

\SubSubSection{Visualization and Analysis} \label{sec:exp-vis}
To provide a better understanding of the tree learning and ZSFA, we visualize the results in several ways. 
First, we illustrate the tree routing results.
In Fig.~\ref{fig:vis1}, we rank the spoof data based on the routing function values $\varphi(\textbf{\textit{x}})$, and provide $8$ examples with responses from the smallest to the largest.
This offers us an intuitive understanding of what are learned at each tree node.
We observe an obvious spoof style transfer: for the first two-layer nodes $N_1$, $N_2$ and $N_3$, the transfer captures the change of general spoof attributes such as image quality and color temperature; for the third-layer tree nodes $N_4$, $N_5$, $N_6$, and $N_7$, the transfer involves more spoof type specific changes.
E.g., $N_7$ transfers from eye portion spoofs to full face $3$D mask spoofs.

Further, Fig.~\ref{fig:vis2} quantitatively analyzes the tree routing distributions of all types of data. 
We utilize two models, Print and Trans.~Mask, to generate the distributions.
It can be observed that live samples are relatively more spread out to $8$ leaf nodes while the spoof attacks are routed to fewer specific leaf nodes.
Two distributions in Fig.~\ref{fig:vis2} (a)\&(b) share similar semantic sub-groups, which demonstrates the success of the proposed method on learning a tree.
E.g., in both models, about half of trans.~mask samples  share the same leaf node as ob.~makeup. 
By comparing two distributions, most testing unknown spoofs in both models are successfully routed to the most similar sub-groups.

\begin{figure}[t!]
\centering
\includegraphics[width=0.9\linewidth]{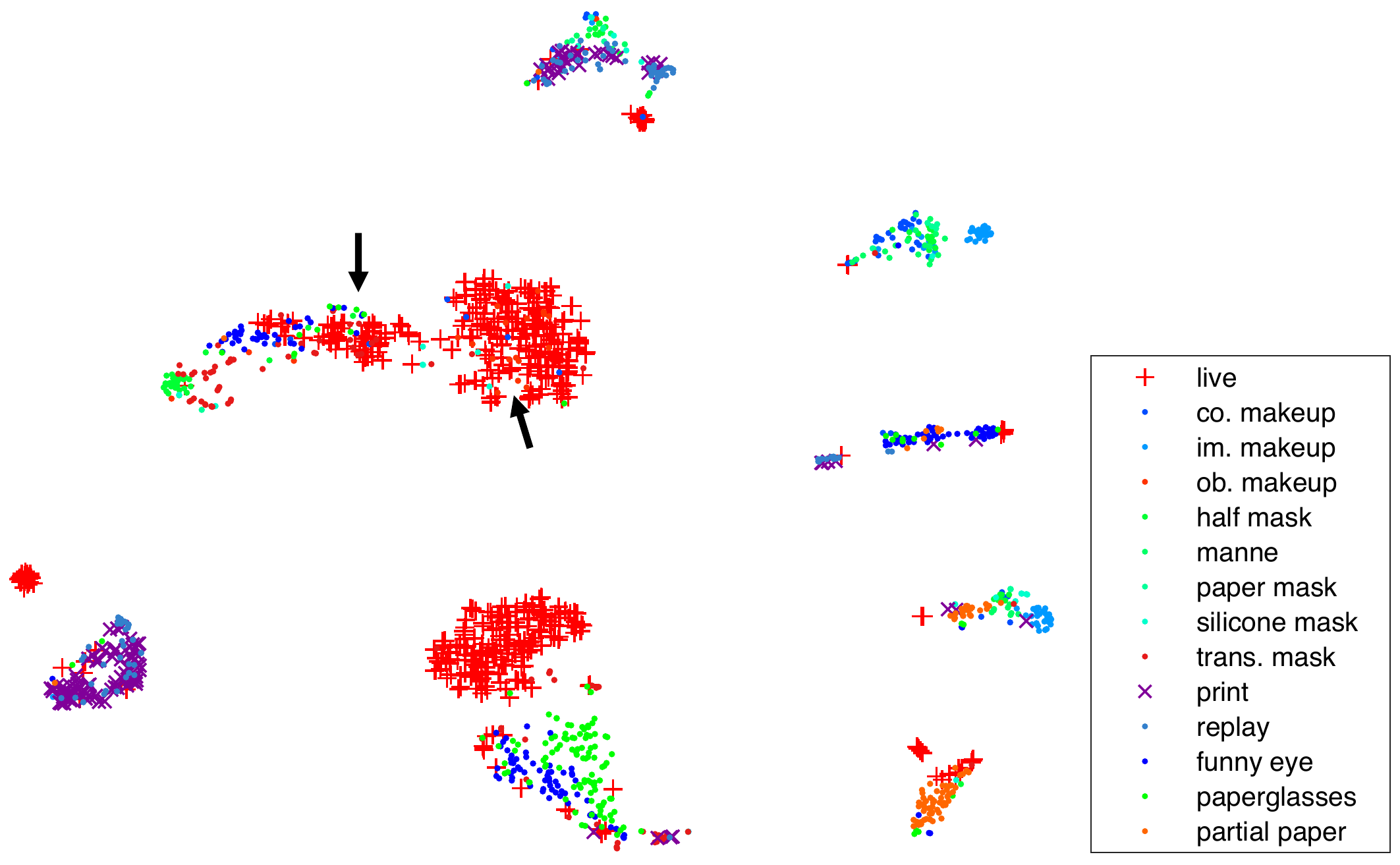}
\vspace{-3mm}
\caption{\small t-SNE Visualization of the DTN leaf features.}
\label{fig:vis3}
\figvspace 
\end{figure}

In addition, we use t-SNE~\cite{maaten2008visualizing} to visualize the feature space of Print model.
The t-SNE is able to project the output of the leaf node $\mathcal{F}(\mathbf{I} \, | \, \theta) \in \mathbb{R}^{32 \times 32 \times 40}$ to $2$D by  preserving the KL divergence distance. 
Fig.~\ref{fig:vis3} shows the features of different types of spoof attacks are well-clustered into $8$ semantic sub-groups even though we don't provide any auxiliary labels.
Based on these sub-groups, the features of unknown print attacks are well lied in the sub-group of replay and silicone mask, and thus are recognized as spoof. 
Moreover, with the visualization, we can explain the performance variation among different spoof attacks, shown in Tab.~\ref{tab:test}. 
Among all, the performance of trans.~mask, funny eye, paper glasses and ob.~makeup are worse than other protocols. 
The feature space shows that the live samples lies much closer to those attacks than others (``$\rightarrow$'' places), and hence it's harder to distinguish them with the live samples. 
This demonstrates the diverse property of different unknown attacks and the necessity of such a wide range evaluation.
\Section{Conclusions}
\label{sec:con}

This paper tackles the zero-shot face antispoofing problem among $13$ types of  spoof attacks. The proposed method leverages a deep tree network to route the unknown attacks to the most proper leaf node for spoof detection. The tree is trained in an unsupervised fashion to find the feature base with the largest variation to split the spoof data. We collect SiW-M that contains more subjects and spoof types than any previous databases. Finally, we experimentally show  superior performance of the proposed method.

\figvspace 
\paragraph{Acknowledgment}
This research is based upon work supported by the Office of the Director of National Intelligence (ODNI), Intelligence Advanced Research Projects Activity (IARPA), via IARPA R\&D Contract No.~$2017$-$17020200004$. The views and conclusions contained herein are those of the authors and should not be interpreted as necessarily representing the official policies or endorsements, either expressed or implied, of the ODNI, IARPA, or the U.S. Government. The U.S. Government is authorized to reproduce and distribute reprints for Governmental purposes notwithstanding any copyright annotation thereon.

{\small
\bibliographystyle{ieee}
\bibliography{egbib}
}

\end{document}